# Multi-Stage Temporal Difference Learning for 2048-like Games

Kun-Hao Yeh[1], I-Chen Wu[1], *Senior Member, IEEE*, Chu-Hsuan Hsueh[1], Chia-Chuan Chang[1], Chao-Chin Liang[1], Han Chiang[1]

*Abstract* —Szubert and Jaśkowski successfully used temporal difference (TD) learning together with n-tuple networks for playing the game 2048. However, we observed a phenomenon that the programs based on TD learning still hardly reach large tiles. In this paper, we propose multi-stage TD (MS-TD) learning, a kind of hierarchical reinforcement learning method, to effectively improve the performance for the rates of reaching large tiles, which are good metrics to analyze the strength of 2048 programs. Our experiments showed significant improvements over the one without using MS-TD learning. Namely, using 3-ply expectimax search, the program with MS-TD learning reached 32768-tiles with a rate of 18.31%, while the one with TD learning did not reach any. After further tuned, our 2048 program reached 32768-tiles with a rate of 31.75% in 10,000 games, and one among these games even reached a 65536-tile, which is the first ever reaching a 65536-tile to our knowledge. In addition, MS-TD learning method can be easily applied to other 2048-like games, such as Threes. Based on MS-TD learning, our experiments for Threes also demonstrated similar performance improvement, where the program with MS-TD learning reached 6144-tiles with a rate of 7.83%, while the one with TD learning only reached 0.45%.

## I. INTRODUCTION

Recently, 2048-like games, single-player stochastic games including 2048[2], 1024[3] and Threes[4], have been very popular over the Internet, especially for 2048. Both 2048 and 1024 were actually originated from Threes. Gabriele Cirulli [14], the author of 2048, claimed his estimation: the aggregated time of playing the game online by players during the first three weeks after release was over 3,000 years. The game is intriguing and even addictive to human players, because it is hard to win the game despite the simple rules. For the same reason, the game also attracted many programmers to develop artificial intelligence (AI) programs to play it. In [21], the authors also thought that the game was an interesting testbed for studying AI methods.

Many methods were proposed to design AI programs for 2048 and Threes in the past. Most commonly used methods were alpha-beta search [10][15][18], a traditional game search method for two-player games, and expectimax search [2][12][18], a common game search method for single-player stochastic games. Recently, Szubert and Jaśkowski [21] proposed *Temporal Difference* (*TD*) *learning* together with n-tuple networks for 2048. They successfully used it to reach a win rate (the rate of reaching 2048-tiles) of 97%, and obtain the average score 100,178 with maximum score 261,526. However, we observed a phenomenon: the TD learning method tends to maximize the average scores, but becomes less motivated to reach large tiles, such as 16384 or 32768, even with expectimax search incorporated.

To cope with this problem, we propose *multi-stage TD* (*MS-TD*) learning, a kind of hierarchical reinforcement learning method. In MS-TD, we separate the training into multiple stages. Our experiments showed significant improvements over the one without using MS-TD learning for 2048 and Threes, especially in the reaching rates of large tiles, which are good metrics to analyze the performance. We consider the 32768-tile reaching rate for 2048 and the 6144-tile reaching rate for Threes.

Our experiments also showed improvements when further incorporating expectimax search. For 2048, the program incorporating MS-TD with 3-ply expectimax search reached 32768-tiles with a probability of 18.31%, while the program with TD learning only did not reach any 32768-tile. After more improvements together with expectimax search, our 2048 program reached 32768-tiles with a probability of 31.75% in 10,000 games. Interestingly, one among these games reached a 65536-tile, the first ever reaching a 65536-tile to our knowledge.

Similarly for Threes, the experiments also showed significant improvement of the Threes program based on MS-TD. Namely, it reached 6144-tiles with a probability of 7.83%, while the program with TD learning only reached 6144-tiles with a much lower probability of 0.45%.

Note that the preliminary version of this paper [32] did not include the following: MS-TD applied to Threes, more splitting strategies for both 2048 and Threes, and a demonstration of combining MS-TD with other techniques to improve 2048 and Threes programs.

This paper is organized as follows. Section II reviews background knowledge. Section III proposes MS-TD learning. Section IV does experiments and analysis for MS-TD learning for 2048, and Section V for Threes. Section VI makes concluding remarks.

---

[1] This work was supported in part by the Ministry of Science and Technology of the Republic of China (Taiwan) under Contracts MOST 104-2221-E-009-127-MY2 and 104-2221-E-009-074-MY2. The authors are with the Department of Computer Science, National Chiao Tung University, Hsinchu 30050, Taiwan. (e-mail of the correspondent: icwu@csie.nctu.edu.tw)

[2] Game 2048. [Online]. Available: http://gabrielecirulli.github.io/2048/
[3] Game 1024. [Online]. Available: http://1024game.org/
[4] Game Threes. [Online]. Available: http://asherv.com/threes

## II. BACKGROUND

This section introduces the rules of 2048 and Threes in Subsection II.A, describes game tree search in Subsection II.B, reviews TD learning in Subsection II.C, and discusses n-tuple networks for 2048 proposed in [21] in Subsection II.D.

### A. 2048 and Threes

The game 2048 can be played on web pages and mobile devices with a 4x4 board, where each cell is either empty or placed with a tile labeled with a value which is a power of two. Let $v$-tile denote a tile with value $v$. Initially, two of 2-tiles or 4-tiles are placed on the board at random. In each turn, the player makes a move and then the game generates a new 2-tile with a probability of 9/10 or 4-tile with a probability of 1/10 on an empty cell chosen at random.

To make a move, the player chooses one of the four directions, up, down, left and right. Upon choosing a direction, all the tiles move in that direction as far as they can until they reach the border or there is already a different-label tile next to it. When sliding a tile, say $v$-tile, if the tile next to it is also a $v$-tile, then the two tiles will be merged into a larger tile, $2v$-tile. At the same time, the player gains $2v$ more points in the score. A move is legal if at least one tile can be moved.

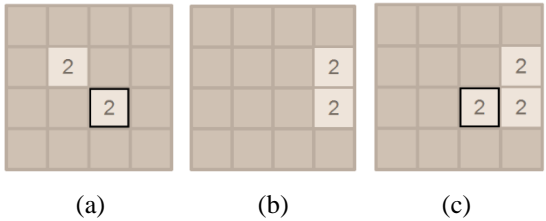

Fig. 1 Examples of 2048 boards.

Consider an example, in which an initial board is shown in Fig. 1 (a). After making a move to right, the board becomes the one shown in Fig. 1 (b). Then, a new 2-tile is randomly generated as shown in Fig. 1 (c). The player repeatedly makes moves in this way.

A game ends when the player cannot make any legal move. The final score is the points accumulated during the game. The objective of the game is to accumulate as many points as possible. The game claims that the player wins when a 2048-tile is created, but still allows players to continue playing optionally.

It is observed that the game 2048 has a phenomenon, also called *survival phenomenon* in this paper. A 2048 game often ends when nearly reaching a large tile, say 16384-tile, due to being blocked by some other smaller large tiles, such as 8192-tiles and 4096-tiles. However, once the game successfully reaches a 16384-tile, most of these smaller large tiles are gone usually. The game survives in the sense that it can usually continue to be played without being blocked by these large tiles for a while, and thus obtain a much higher score. So, the score gap between the games reaching 16384-tiles and the games not reaching the tile is usually high. From the above observations, the reaching rates of large tiles are a good metric to stand for playing strength. In this paper, the 32768-tile reaching rate is used.

In the game Threes, the board size is 4x4 as well. When the game starts, every cell either is empty or has one $v$-tile, where $v$ is 1, 2 or 3. The rule of tile moving is slightly different from 2048. The sliding distance is at most one. In the rules of merging, a 1-tile and a 2-tile can be merged into a 3-tile, and two $v$-tiles can be merged into a $2v$-tile like the game 2048, where $2 < v < 6144$. Note that 6144-tiles cannot be merged anymore. The rules of generating new tiles are much more complex than those for 2048 [8]. Initially, there is a bag of 12 tiles composed of equal amount of 1, 2, and 3-tiles. A *normal random tile* is randomly selected from the bag until the bag is empty and then refilled. Let $v_{max}$-tile denote the tile labeled with the largest value $v_{max}$ on the current board. If $v_{max} \geq 48$, a new tile could be a *bonus random tile*, namely a $v$-tile with $v > 3$, which is generated with a probability of 1/21 while other tiles from the bag are generated with 20/21. A bonus random tile ranges from a 6-tile to a $v$-tile, where $v = v_{max} * (1/8)$, with equal probability. For example, if $v_{max} = 192$, then the bonus random tile is one of 6-tile, 12-tile and 24-tile with equal probabilities.

A Threes game ends similarly when the player is no longer able to make any legal move. The final score is the sum of scores of all $v$-tiles with $v \geq 3$, using the formula: $\text{score} = 3^{\log_2\left(\frac{v}{3}\right)+1}$. The objective of this game is to achieve as much score as possible. The game does not define a win like 2048. The above survival phenomenon also exists in Threes. In this paper, the 6144-tile reaching rate is used for Threes.

In a 2048-bot tournament [22], contests for both 2048 and Threes were held separately. For the 2048 contest, all the 2048-bot participants played 100 games. Their performances were graded in a formula (described in [22]) from the following criteria, the win rates, the average scores, the maximum scores, and the reaching rates of large tiles. The Threes contest was similar except that reaching 192-tiles was defined as a win.

### B. Game Tree Search

A common game tree search algorithm used in 2048 programs is expectimax search [2][12][18]. Like most game tree search, the leaves are evaluated with values calculated by heuristic functions. An expectimax search tree contains two different kinds of nodes, max nodes and chance nodes. At a max node, its value is the highest value of its children, if any. At a chance node, its value is the expected value of its children, if any, weighted by the probabilities of children. An expectimax search tree is illustrated in Fig. 2.

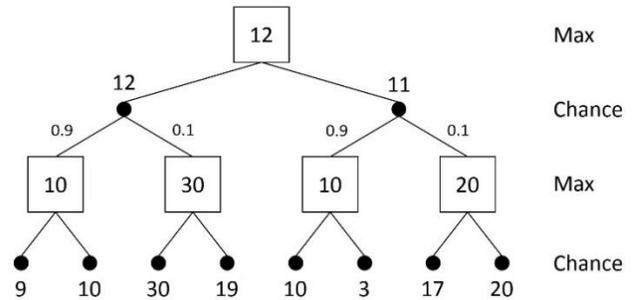

Fig. 2. An expectimax search tree.

Several kinds of features were used in heuristic functions for 2048 programs that use game tree search [13][30]. The first is the monotonicity of a board. Most high-ranked players tend to play 2048 with tiles arranged decreasingly in several ways, as described in [18]. The second is the number of empty tiles on a board. The more empty tiles, the less likely for the game to end in a few steps. The third is the number of pairs of mergeable tiles, since it is a measurement of the ability to create empty tiles by merging tiles. A pair of mergeable tiles are two neighboring tiles which can be merged, e.g., two tiles with the same labels in 2048.

A transposition table is a technique to avoid searching the same positions redundantly and therefore speed up the search. One common implementation is based on Zobrist hashing [33]. In Zobrist hashing, for each cell, each kind of possible tile is assigned a unique random number as a key. When looking up table, the hash value used to access the transposition table is calculated by doing an exclusive-or operation on the 16 keys for all 4x4 cells.

### C. Temporal Difference (TD) Learning

*Reinforcement learning* is an important technique in training an agent to learn how to respond to a given environment [20]. A *Markov decision process* (*MDP*) is a model commonly used in reinforcement learning, modeling the problems in which an agent interacts with the given environment through a sequence of actions according to the change of the state and the rewards, if any. In terms of MDP, an AI program for a 2048-like game thus can be regarded as such an agent, which makes actions (legal moves) on board states and gets points as rewards.

*Temporal difference* (*TD*) *learning* [19][20], a kind of reinforcement learning, is a method for adjusting state values from the subsequent evaluations. This method has been applied to several computer games such as Backgammon [23], Checkers [16], Chess [4], Shogi [5], Go [17], Connect6 [31][32], Othello [9][11][28], Connect four [1][24][25] and Chinese Chess [27]. Among the above, TD learning was demonstrated to improve some world class game-playing programs, e.g., Chinook [16], and TD-Gammon [23]. Since 2048-like games can be easily modeled as MDP, TD learning can be naturally applied to AI programs for these games.

In TD(0), the value function $V(s)$ is used to approximate the expected return of a state $s$. The error between states $s_t$ and $s_{t+1}$ is $\delta_t = r_t + V(s_{t+1}) - V(s_t)$, where $r_t$ is the reward at turn $t$. The value of $V(s_t)$ in TD(0) is expected to be adjusted by the following value difference $\Delta V(s_t)$,

$$\Delta V(s_t) = \alpha \delta_t = \alpha\bigl(r_t + V(s_{t+1}) - V(s_t)\bigr) \quad (1)$$

where $\alpha$ is a step-size parameter to control the learning rate. Note that in [20] $V(s_{t+1})$ is weighted by a discount factor, which is ignored in this paper for simplicity. For the general TD($\lambda$) as in [19][20][31], we adopted the forward view of value difference as mentioned (c.f. Section 7.2 of [20]):

$$\Delta V(s_t) = \alpha\left(R_t^\lambda - V(s_t)\right), \quad (2)$$

$$R_t^\lambda = (1-\lambda)\sum_{n=1}^{T-t-1}\lambda^{n-1}R_t^n + \lambda^{T-t-1}R_t^{T-t}, \quad (3)$$

$$R_t^n = \sum_{k=0}^{n-1} r_{t+k} + V(s_{t+n}), \quad (4)$$

where $T$ is the ending time step. In this paper, TD(0) is investigated, unless specified.

In most applications, the evaluation function of states $V(s)$ can be viewed as a function of features, such as the monotonicity, the number of empty tiles, and the number of mergeable tiles for 2048 [13], mentioned in Subsection II.B. The function is usually modified into a linear combination of features [31] for TD learning, that is, $V(s) = \varphi(s) \cdot \theta$, where $\varphi(s)$ denotes a vector of feature occurrences in $s$, and $\theta$ denotes a vector of feature weights.

In order to correct the value $V(s_t)$ by the difference $\Delta V(s_t)$, we adjust the feature weights $\theta$ by a difference $\Delta\theta$ based on the gradient $\nabla_\theta V(s_t)$, which is $\varphi(s_t)$ for linear TD(0) learning. Thus, the difference $\Delta\theta$ is

$$\Delta\theta = \Delta V(s_t)\varphi(s_t) = \alpha\delta_t\varphi(s_t). \quad (5)$$

In [21], Szubert and Jaśkowski proposed TD learning for 2048. A *transition* from turn $t$ to $t+1$ is illustrated in Fig. 3. They also proposed three kinds of methods to evaluate values for training and learning as follows.

1. Evaluate actions. This method is to evaluate the function $Q(s, a)$, which stands for the expected values of taking an action $a$ on a state $s$. For 2048, an action $a$ is one of the four directions, up, down, left, and right. This is so-called *Q-learning*.

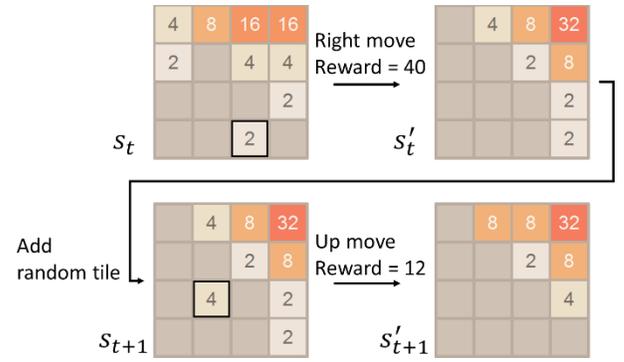

Fig. 3. Transitions of board states.

2. Evaluate states to play. This method is to evaluate the value function $V(s_t)$ on state $s_t$, the player to move.
3. Evaluate states after an action. This method is to evaluate the value function $V(s_t')$ on state $s_t'$, a state after an action, also called an *afterstate* in [21].

In [21], their experiments showed that the third method clearly outperformed the other two. In the rest of this paper, we only consider the third, evaluating afterstates, and let states refer to afterstates for simplicity.

### D. N-Tuple Network

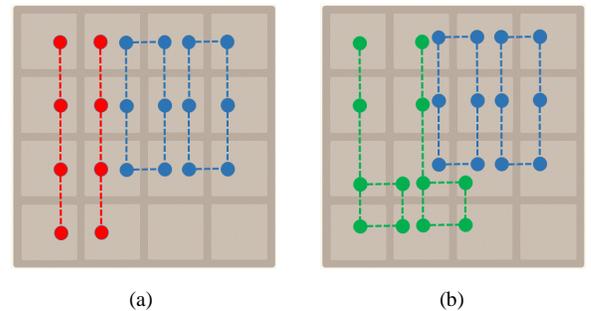

(a)        (b)

Fig. 4. (a) Tuples used in [21] and (b) tuples used in this paper.

In [21], they also proposed to use n-tuple networks for TD learning in 2048. In fact, n-tuple networks were also successfully applied to other applications such as Othello [9][11][28] and Connect four [1][24][25]. An n-tuple network consists of $m$ $n_i$-tuples, where $n_i$ is the size of the $i$-th tuple. As shown in Fig. 4 (a), one 4-tuple covers four cells marked in red dots and one 6-tuple covers six cells marked in blue dots. Each tuple contributes a large number of features, each for one distinct occurrence of tiles on the covered cells. For example, the leftmost 4-tuple in Fig. 4 (a) includes $16^4$ features, assuming that a cell has 16 occurrences, empty or 2-tile to $2^{15}$-tile.

The output of a network is a linear summation of feature weights for all occurring features. For each tuple, since one and only one feature occurs at a time, the feature weight can be easily accessed by looking up a table. If an n-tuple network includes $m$ different tuples, we need $m$ lookups.

In [21], they used the tuples shown in Fig. 4 (a) as well as all of their rotated and mirrored tuples. All the rotated and mirrored tuples can share the same feature weights. Thus, the total number of features was roughly 2x15$^4$+2x15$^6$, about 23 millions. Their experiments in [21] showed an average score of 100,178 and a maximum score of 261,526. In this paper, we use the n-tuple network, as shown in Fig. 4 (b), by changing 4-tuples (1x4 lines in red) in Fig. 4 (a) to 6-tuples in knife-shaped ones in Fig. 4 (b). Apparently, the new 6-tuples cover all the original 4-tuples while still not covering the 6-tuples in 2x3. The number of features in total increases to 4x16$^6$, about triple[5] of the one used in [21] only.

## III. MULTI-STAGE TD LEARNING ALGORITHM

From above, TD learning is intrinsically to train and learn to obtain average (or expected) scores as high as possible. The experiments in [21] also demonstrated this. However, TD learning does not necessarily lead to other criteria such as high maximum scores, or the reaching rates of large tiles, though it does often.

From the experience of playing 2048, we observed that it was hard to push to higher maximum scores, or raise the reaching rates of 32768-tiles based on the original TD learning, even with expectimax search. However, obtaining the maximum scores as well as reaching large tiles is a kind of goal or achievement for most players, and was also one of the criteria of the 2048-bot tournament [22] as described in Subsection II.A.

In order to address this issue, we propose multi-stage TD (MS-TD) learning for 2048-like games. In this method, we divide the learning process into multiple stages, each of which has its own learning agent and subgoal, e.g., reaching 8192-tiles or 16384-tiles, as in hierarchical reinforcement learning (HRL) [3]. In HRL, different agents use their own sets of feature weights. The concept of using different feature weights in multiple stages was also mentioned in the work [6] to evaluate game states and select different features according to different game stages, but not for reinforcement learning. In our method, different feature weights are trained and used in different stages to improve the original TD learning.

The method using a simple 3-stage strategy for 2048 is illustrated as follows. We divide the process into three stages with two splitting times, marked as $T_{16k}$ and $T_{16+8k}$, in games. $T_{16k}$ denotes the first time when a 16384-tile is created on the board, and $T_{16+8k}$ denotes the first time when both 16384-tile and 8192-tile are created.

The learning process in the three stages is described respectively as follows. In the first stage, use TD learning to train the feature weights for millions of training games, until the learning is *saturated*, namely the average scores in every 1,000 games gradually stabilize without further significant improvement. Usually, we train a certain number of games, say 5 million games, to saturation. The set of trained feature weights are called *Stage-1 feature weights*. After saturation, the program keeps playing games to collect a number of boards, say 100,000 samples, at $T_{16k}$, and their scores, which become the initial boards and scores for the training in the next stage. Note that Stage-1 feature weights remain unchanged during collection.

In the second stage, use TD learning to train another new set of feature weights starting from these collected boards in the first stage. The collected boards are repeatedly used in a round-robin manner until millions of training games are trained. After finishing the training, the set of trained feature weights, called *Stage-2 feature weights*, are saved. Then, collect boards at $T_{16+8k}$ for the third stage in a similar way as the first stage.

In the third stage, use TD learning to train another new set of feature weights starting from those collected boards in the second stage. Again, the set of trained feature weights, called *Stage-3 feature weights*, are saved. For simplicity, feature weights are all initialized to be zero.

When playing a game, we also divide the process into three stages in the following way.

1. Before $T_{16k}$, use Stage-1 feature weights to play.
2. After $T_{16k}$ and before $T_{16+8k}$, use Stage-2 feature weights to play.
3. After $T_{16+8k}$, use Stage-3 feature weights to play.

The idea behind using more stages is to make learning more accurate for all states during the second and the third stages, based on the following observation. The feature weights learned from the first stage (the same as the original TD learning) tend to perform well in the first stage, but may not in the rest of stages. In these stages, large tiles such as 16384-tile and 8192-tile increase the difficulty of playing the game, since these large tiles are more difficult to be merged into a larger one. Therefore, the feature weights may not accurately reflect the expected scores with the difficulty. Hence, we use different sets of feature weights in different stages in order to make the feature weights reflect the difficulty in the expected scores.

The effectiveness of MS-TD learning is justified in the experiments in Section IV and Section V, with significant improvements for 2048 and Threes.

## IV. EXPERIMENTS FOR 2048

In this section, experiments are done to analyze the performances of MS-TD learning for 2048, on machines equipped with AMD Opteron 6174 x4, 128GB RAM, Linux. First, in Subsection IV.A, we modify the n-tuple network from that by Szubert and Jaśkowski [21] for subsequent experiments. Second, experiments for the above simple 3-stage strategy (as illustrated in Section III) are done and analyzed in Subsection IV.B, and expectimax search for the strategy is described in Subsection IV.C. More splitting strategies are experimented and analyzed in Subsection IV.D. Subsection IV.E discusses more methods to further improve MS-TD learning.

---

[5] Our n-tuple networks contain 32768-tiles, while those in [21] do not.

### A. New N-Tuple Network and Feature

In our experiments, we used (i) the tuples shown in Fig. 4 (b) as well as all of their rotated and mirrored tuples, and (ii) large-tile features, which is a set of features representing all the combinations of large tiles, namely $v$-tiles, where $v \geq 2048$. Namely, the large-tile feature for one game state is indexed by the tuple: $(n_{2048}, n_{4096}, n_{8k}, n_{16k}, n_{32k})$, denoting the numbers of 2048-tile(s), 4096-tile(s), 8192-tile(s), 16384-tile(s) and 32768-tile(s), respectively. The features in (ii) were used to indicate difficulty due to large tiles. In our implementation, a state is evaluated as the sum of all occurring features' weights. We tried all combinations for both (i) and (ii). The n-tuple network with both (i) and (ii) outperformed others in terms of both average and maximum scores, as shown in Fig. 5 and Fig. 6 respectively. Note that the step sizes for both were set to $\alpha = 0.0025$, the same as that in [21]. In these figures, the number of training games is 2 millions and average/maximum scores in y-axis are sampled every 1,000 games. For simplicity of analysis, we use the n-tuple network as shown in Fig. 4 (b) and additional features described above in the rest of this paper.

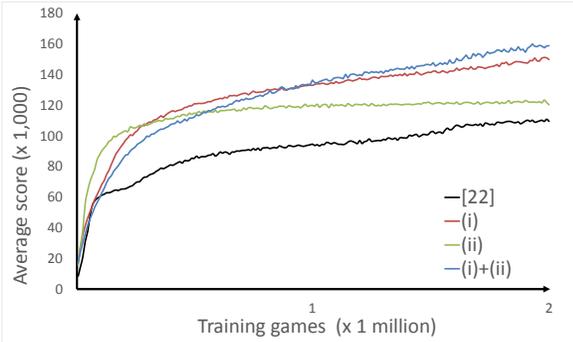

Fig. 5. Average scores in 2048 for TD learning with all combinations for both (i) and (ii).

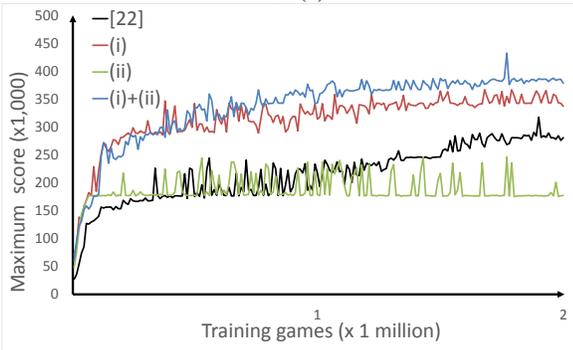

Fig. 6. Maximum scores in 2048 for TD learning with all combinations for both (i) and (ii).

### B. MS-TD Learning for the Simple 3-Stage Strategy

In the experiment for MS-TD learning, 5 million training games were run in each stage, and 100,000 game boards were collected for the next stage. All feature weights of each stage were initialized to zero.

Fig. 7 (below) shows the learning curves of average scores (sampled every 100,000 games) in the three stages. The curve for Stage 1 is depicted like Fig. 5. For fair comparison, we used the following method to depict other curves. For illustration, we first consider the curve for Stage 2. In Stage 2, we trained games starting from the splitting points ($T_{16k}$) of the 100,000 games collected from Stage 1. The scores for games trained in Stage 2 included the scores accumulated up to $T_{16k}$ in Stage 1. To fairly compare the performances of Stage-1 feature weights with Stage-2 feature weights, we calculated the average score of the 100,000 collected games (in Stage 1), which included all scores after $T_{16k}$. The average score, shown as a (blue) dashed line in Fig. 7, is called the *normalized score of Stage 1 with respect to Stage 2*, or called the *normalized 1-2 score*. The comparison between the curve of Stage 2 and the normalized score is fair in the sense that both include all the scores accumulated before and after $T_{16k}$. The curve for Stage 3 and the normalized 2-3 score were derived similarly.

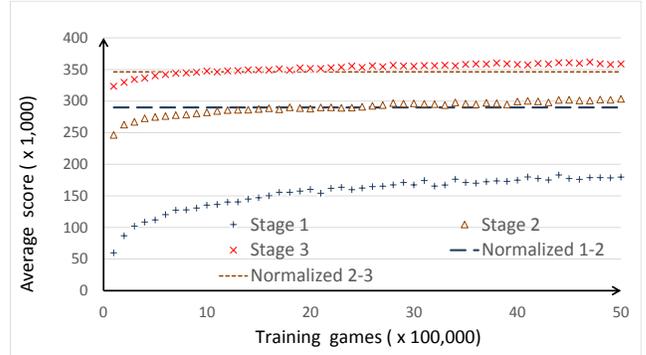

Fig. 7. Average scores of different stages in 2048 for MS-TD learning.

In Fig. 7, the learning curve of Stage 2 grows higher than normalized 1-2 after about two million games. Similarly, the learning curve of Stage 3 grows higher than normalized 2-3 after about 1.5 million games. The results showed that the learning in the second stage did slightly improve over the first stage in terms of average scores, and the same for the third over the second.

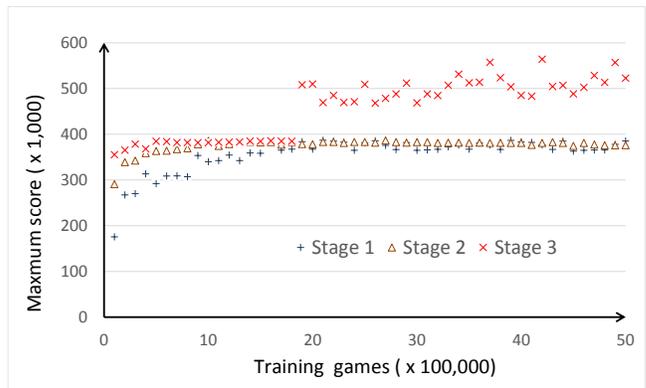

Fig. 8. Maximum scores of different stages in 2048 for MS-TD learning.

Fig. 8 shows the curves of maximum scores (sampled every 100,000 games) in the three stages. For fair comparison, we obtained maximum scores in the second and the third stages as follows. For example, in the experiments, if 30% of games in the first stage can reach the second, the maximum scores for Stage 2 were only retrieved from 30,000 randomly selected games of every 100,000 games, namely the first 30,000 games in this paper. The maximum scores for the third stage were obtained similarly. In this figure, the curve for the third stage does go up to 500,000 often, which actually indicates to reach 32768-tiles.

In contrast, the curves for the first and the second stages rarely reach 500,000 points. This demonstrated that maximum scores can be significantly improved by using MS-TD learning.

*C. MS-TD Learning Together with Expectimax Search*

Expectimax search fits afterstates evaluation well. As described in Subsection II.B, max nodes are the states, where players are allowed to move, and chance nodes are the afterstates, where new tiles are generated after moves. For TD learning, the learned afterstate values can be used as the heuristic values of leaves. Thus, choosing the maximum afterstate values can be viewed as 1-ply expectimax search. Fig. 2 shows an example of 2-ply expectimax search.

| Reaching rate | 1-ply | 2-ply | 3-ply |
|---|---|---|---|
| 2048 | 96.88% ($\pm$0.34%) | 99.84% ($\pm$0.08%) | 99.97% ($\pm$0.03%) |
| 4096 | 90.99% ($\pm$0.56%) | 99.51% ($\pm$0.14%) | 99.86% ($\pm$0.07%) |
| 8192 | 71.46% ($\pm$0.89%) | 97.02% ($\pm$0.33%) | 99.04% ($\pm$0.19%) |
| 16384 | 26.33% ($\pm$0.86%) | 79.86% ($\pm$0.79%) | 90.95% ($\pm$0.56%) |
| 32768 | 0.00% ($\pm$0.00%) | 0.00% ($\pm$0.00%) | 0.00% ($\pm$0.00%) |
| Maximum score | 371,595 ($\pm$15,086) | 374,675 ($\pm$2,274) | 374,458 ($\pm$3,344) |
| Average score | 159,565 ($\pm$6,843) | 290,381 ($\pm$6,066) | 321,496 ($\pm$2,859) |
| Speed (moves/sec) | 280,882 ($\pm$22,334) | 13,861 ($\pm$2,102) | 1,256 ($\pm$118) |

Table 1: Results of 10,000 games for expectimax search using values learned from TD learning in 2048.

| Reaching rate | 1-ply | 2-ply | 3-ply |
|---|---|---|---|
| 2048 | 96.60% ($\pm$0.36%) | 99.81% ($\pm$0.09%) | 99.99% ($\pm$0.02%) |
| 4096 | 90.05% ($\pm$0.59%) | 99.47% ($\pm$0.14%) | 99.87% ($\pm$0.07%) |
| 8192 | 67.48% ($\pm$0.92%) | 96.24% ($\pm$0.37%) | 98.79% ($\pm$0.21%) |
| 16384 | 17.87% ($\pm$0.75%) | 74.36% ($\pm$0.86%) | 87.68% ($\pm$0.64%) |
| 32768 | 0.06% ($\pm$0.05%) | 5.50% ($\pm$0.45%) | 13.78% ($\pm$0.68%) |
| Maximum score | 431,847 ($\pm$78,795) | 554,090 ($\pm$57,640) | 599,645 ($\pm$33,701) |
| Average score | 143,958 ($\pm$5,620) | 301,280 ($\pm$6,996) | 350,394 ($\pm$8,579) |
| Speed (moves/sec) | 213,729 ($\pm$51,528) | 13,884 ($\pm$2,652) | 1,257 ($\pm$144) |

Table 2: Results of 10,000 games for expectimax search using values learned from 3-stage MS-TD learning in 2048.

Table 1 shows the results of running 10,000 games for the original TD learning for 1-ply to 3-ply expectimax search, respectively, while Table 2 shows those for MS-TD learning with three stages. Maximum scores in the two tables were derived as follows. First, calculate the maximum score for every 100 games. Then, derive the average score of these maximum scores. Besides, all the data in the two tables include errors with 95% confidence intervals. Speeds are the total numbers of moves divided by the total time of generating moves in 10,000 games. For fairer comparison, we ran 15 million training games for TD learning, while running 5 million training games for each of the three stages of MS-TD learning.

From the two tables, when incorporating expectimax search, the performance for MS-TD learning was clearly improved in terms of the 32768-tile reaching rates and the maximum scores. Particularly, with 3-ply search, MS-TD learning significantly improved the maximum score from 374,458 to 599,645, and the reaching rate of 32768-tiles from 0% to 13.78%. Generally, the performance was better for deeper search, but the computation time was longer.

One may notice that for MS-TD learning the reaching rates for smaller $v$-tiles, where $v \leq 16384$, were lower than those for the original TD learning. The reason is: before $T_{16k}$, the original TD learning trained 15 million games, much larger than 5 million training games by MS-TD learning. In addition, the speed for the program trained from the original TD learning was slightly faster than that from MS-TD, since MS-TD used more tables of feature weights that incurred some overhead.

*D. More Splitting Strategies*

In this subsection, we investigate the issue of splitting strategies for MS-TD learning. Let $T_{8k}$ be the first time when an 8192-tile is created, $T_{16+8+4k}$ the first time when a 16384-tile, an 8192-tile and a 4096-tile are created, and similarly for $T_{16+8+4+2k}$ and $T_{16+8+4+2+1k}$. We tried the following four more splitting strategies.

1. Split into four stages at times $T_{8k}$, $T_{16k}$, and $T_{16+8k}$.
2. Split into four stages at times $T_{16k}$, $T_{16+8k}$, and $T_{16+8+4k}$.
3. Split into five stages at times $T_{16k}$, $T_{16+8k}$, $T_{16+8+4k}$, and $T_{16+8+4+2k}$.
4. Split into six stages at times $T_{16k}$, $T_{16+8k}$, $T_{16+8+4k}$, $T_{16+8+4+2k}$, and $T_{16+8+4+2+1k}$.

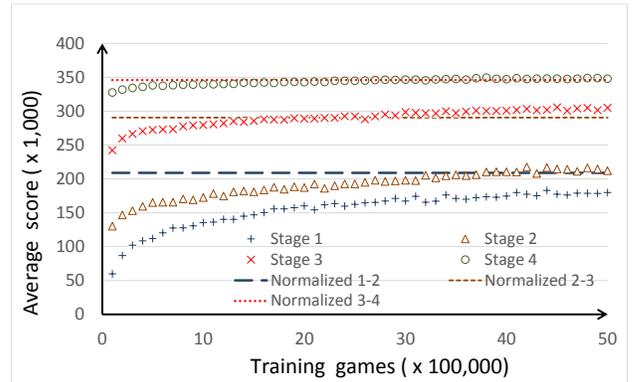

Fig. 9. Average scores in 2048 for Strategy 1.

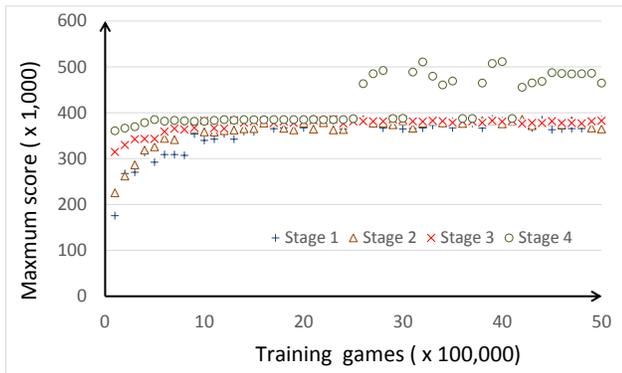

Fig. 10. Maximum scores in 2048 for Strategy 1.

Strategy 1 adds one more splitting time at $T_{8k}$ to the above simple 3-stage strategy. Fig. 9 and Fig. 10 show its learning curves for the average scores and maximum scores, respectively. Similarly, both show the improvement, especially for the maximum scores. However, when compared with Fig. 7 and Fig. 8, Fig. 9 and Fig. 10 show that Strategy 1 did not perform better, that is, adding one more splitting at $T_{8k}$ did not help. Thus, the other three strategies did not split at $T_{8k}$.

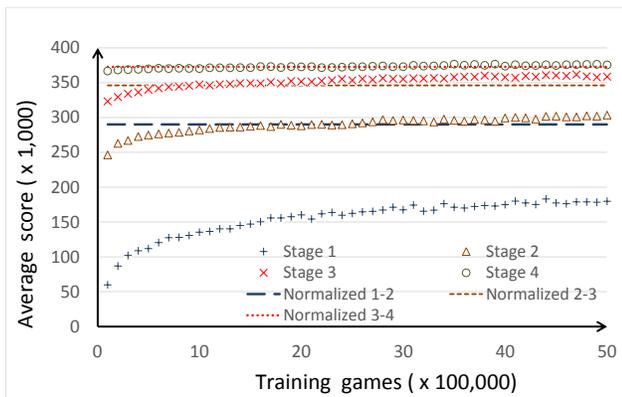

Fig. 11. Average scores in 2048 for Strategy 2.

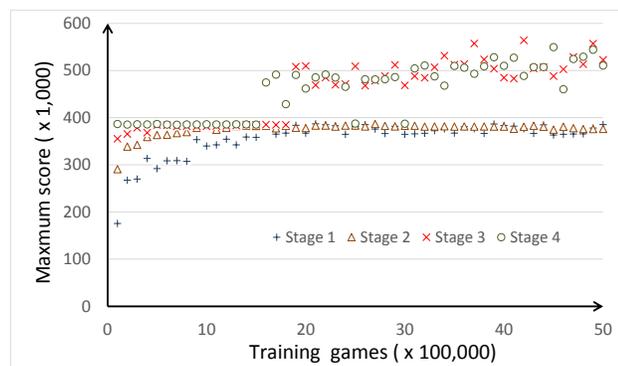

Fig. 12. Maximum scores in 2048 for Strategy 2.

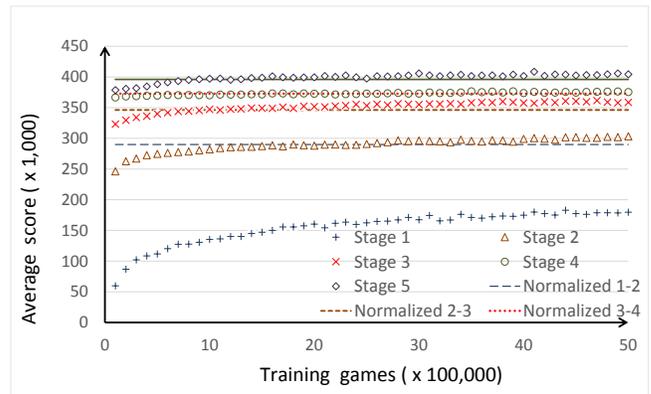

Fig. 13. Average scores in 2048 for Strategy 3.

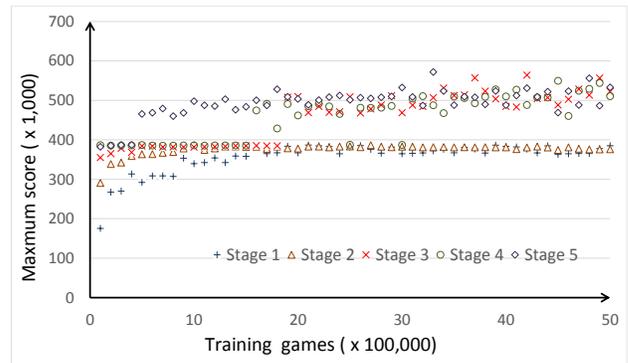

Fig. 14. Maximum scores in 2048 for Strategy 3.

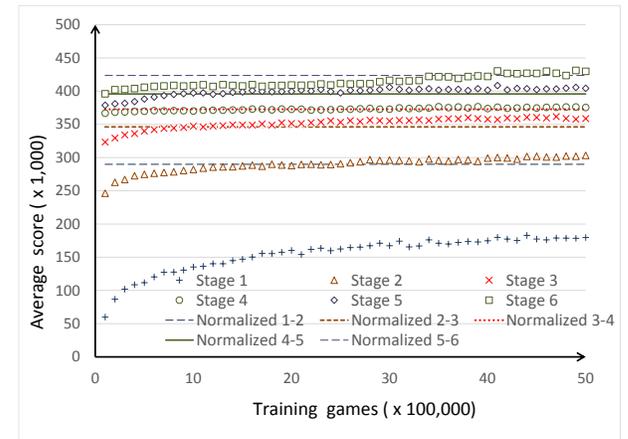

Fig. 15. Average scores in 2048 for Strategy 4.

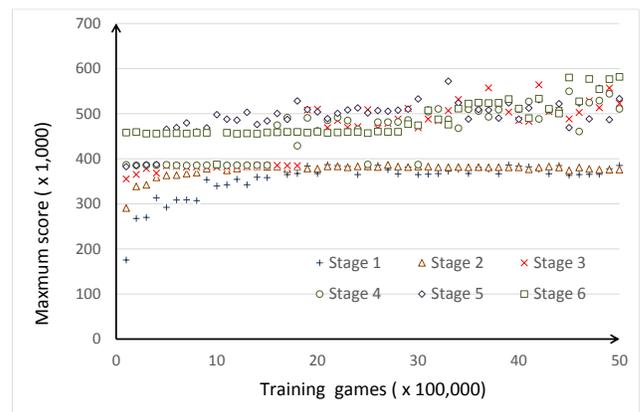

Fig. 16. Maximum scores in 2048 for Strategy 4.

Strategy 2 adds one more at $T_{16+8+4k}$ to the above simple 3-stage strategy, Strategy 3 adds one more at $T_{16+8+4+2k}$ to Strategy 2, and Strategy 4 adds one more at $T_{16+8+4+2+1k}$ to Strategy 3. For the three strategies, the average scores are shown in Fig. 11, Fig. 13 and Fig. 15, respectively, and the maximum scores are shown in Fig. 12, Fig. 14 and Fig. 16, respectively. These figures show that MS-TD learning indeed improved the performances. Particularly, the average scores in Strategy 4 were close to 450,000. And the maximum scores in Strategy 3 and Strategy 4 reached 500,000 very often and some were close to 600,000.

| Reaching rate | Strategy 1 | Strategy 2 | Strategy 3 | Strategy 4 |
|---|---|---|---|---|
| 16384 | 87.23% (±0.65%) | 87.68% (±0.64%) | 87.68% (±0.64%) | 87.68% (±0.64%) |
| 32768 | 14.15% (±0.68%) | 15.30% (±0.71%) | 18.31% (±0.76%) | 16.82% (±0.73%) |
| Maximum score | 594,716 (±51,886) | 579,996 (±40,583) | 614,204 (±10,053) | 610,011 (±13,720) |
| Average score | 349,996 (±3,433) | 350,150 (±9,097) | 361,395 (±9,892) | 364,438 (±9,448) |
| Speed (moves/sec) | 1,152 (±93) | 1,165 (±114) | 1,106 (±106) | 1,078 (±104) |

Table 3: Results of 10,000 games for 3-ply expectimax search using values learned from MS-TD learning with Strategies 1 to 4 in 2048.

Table 3 shows the experiments of playing 10,000 games using 3-ply expectimax search for the above four strategies. Note that the reaching rates of 2048-tile to 8192-tile are not shown since they are all the same. First, in Strategy 1, adding $T_{8k}$ did not improve much in 3-ply expectimax search either. Second, Strategy 4 was the best in terms of average score, while Strategy 3 was the best in terms of maximum score and 32768-tile reaching rate. The 32768-tile reaching rate of Strategy 3 went up to 18.31% but became 16.82% in Strategy 4. This showed that the benefit of splitting extra stage was diminishing. In any case, the results showed that the 32768-tile reaching rates of using MS-TD learning were much higher than that for TD, which was 0% in Table 1. To sum up, the version with Strategy 3 performed the best among all the splitting strategies.

*E. Further Improvements*

In general, we can further improve the performance in the following ways. (a) Add more features. (b) Adjust the step-size parameter $\alpha$ to a smaller value in order to make MS-TD learning more accurate. (c) Use TD($\lambda$) instead of TD(0) for each stage.

In (a), we added features, including the number of empty tiles, the number of distinct tiles, the number of the pairs of mergeable tiles and the number of pairs of neighboring cells with $v$-tile and $2v$-tile. Again, a state is evaluated as the sum of all features' weights. In (b), we adjusted $\alpha = 0.00025$ after the improvement is saturated for $\alpha = 0.0025$. We consider the learning is saturated if the average scores of 1,000 games do not make significant improvement as described in Section III. Finally in (c), we used a variant of TD($\lambda$), $\lambda = 0.5$, and consider only up to the five-step returns (from $s_{t+1}$ to $s_{t+5}$) instead of all returns (from $s_{t+1}$ to $s_T$):

$$R_t^\lambda = 0.5R_t^1 + 0.25R_t^2 + 0.125R_t^3 + 0.0625R_t^4 + 0.0625R_t^5. \quad (6)$$

In Formula (6), the sum of the weights of $n$-step returns ($n$ =1 to 5) is still 1. We used the variant without eligibility traces [19][20] for the following reasons. First, using eligibility traces needs extra effort to deal with the update of sparse feature vectors, especially for those n-tuple networks that represent millions of features. Second, it often takes thousands of moves to finish a game, thus updating from all returns is hence less efficient. For the above reasons, we simply recorded all the state sequences for whole games, and performed updates after the games finished. We leave it open for better variants of TD($\lambda$) [27] [29].

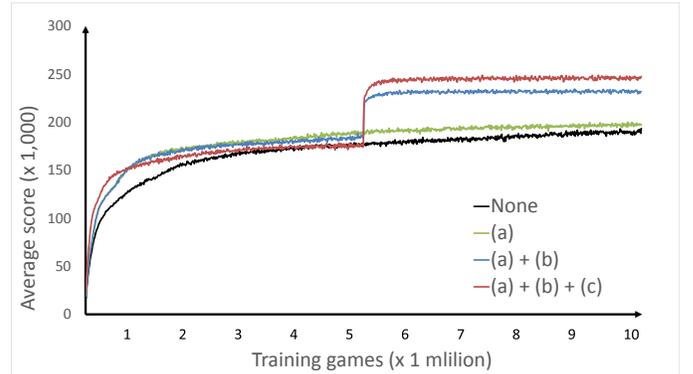

Fig. 17. Improvements for 2048 TD learning with (a), (b), and (c) incrementally applied to the first stage.

We experimented incrementally with the first stage training; that is, added (a), (b), and then (c) one at a time to demonstrate the improvement with these three techniques. These improvements are shown in Fig. 17, and can be summarized as follows. First, (a) improved the training in terms of average score. Second, while (a) was applied, (b) helped to raise the average score significantly. Third, (c) helped reach to a higher average score after (b) was applied. Hence, we applied Strategy 3 (splitting into 5 stages) MS-TD to the above three techniques along with some other fine tunes, and experimented with 3-ply expectimax search. We obtained the following results for this new version in 10,000 games: a 32768-tile reaching rate of 31.75%, the average score 443,526, the maximum score 793,835, and a speed of 500 moves/sec. Note that the maximum score here is the average of maximum scores as described in Subsection IV.C. Interestingly, one among these games even reached a 65536-tile, the first ever reaching a 65536-tile to our knowledge. This version is available openly at [26].

In the past, the one developed by nneonneo and xificurk [13] obtained competitive results to ours. We ran their program for 2,000 games and obtained a 32768-tile reaching rate of 28.25%, the average score 432,557, and the maximum score 814,759 (calculated in the same way as above). However, since their program heavily relied on deep search with tuned heuristics, it ran slowly, about 4 moves/sec, 125 times slower than ours.

## V. EXPERIMENTS FOR THREES

The experiments for Threes were similar to those for 2048. We used the same n-tuple network in Fig. 4 (b) and the features including those mentioned in Section IV.A, and some specific features for Threes, such as the hint tiles. Experiments for a simple 3-stage strategy splitting at $T_{1536}$ and $T_{3072}$ are shown and analyzed in Subsection V.A. Then, more splitting strategies and further improvements are experimented and analyzed in Subsection V.B.

### A. MS-TD Learning for a Simple 3-Stage Strategy

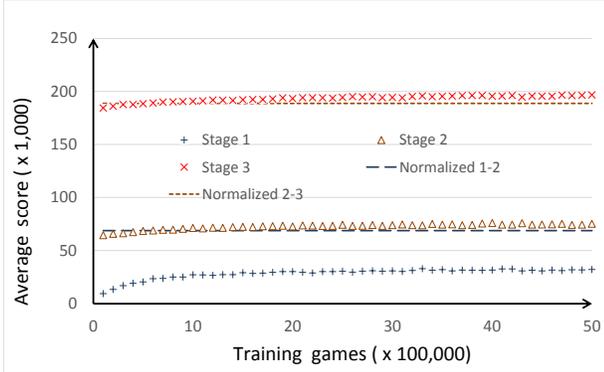

Fig. 18. Average scores in Threes for MS-TD learning.

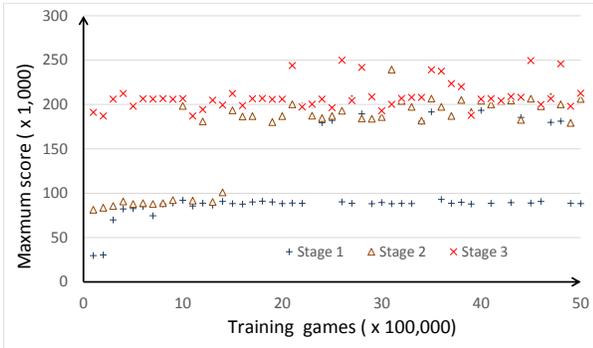

Fig. 19. Maximum scores in Threes for MS-TD learning.

In our experiments for Threes, similar to those for 2048, 5 million training games were run in each stage, and average scores and maximum scores were sampled every 100,000 games. Fig. 18 and Fig. 19 depict the learning curves of average scores and maximum scores in the three stages, respectively. The curves are depicted in the same way as those described in Subsection IV.B. Both figures show that the learning in Stage 2 and Stage 3 did improve, especially for maximum scores. In Fig. 19, the curve for Stage 3 does appear more often around 250,000 than Stage 1 and Stage 2.

Table 4 (below) shows the results of running 10,000 games for the original TD learning together with 1-ply to 3-ply expectimax search, respectively, while Table 5 shows those for MS-TD learning with the simple 3-stage strategy. For simplicity, both tables for reaching rates only include those of large $v$-tiles, namely 3072-tile and 6144-tile.

The results showed that the performance of MS-TD learning clearly outperformed that of the original TD learning. Especially, when using 3-ply expectimax search, the one with MS-TD learning significantly improved the 6144-tile reaching rates from 0.45% to 7.83%. For this one, its average score and maximum score were 220,393 and 700,181, respectively.

Similar to the results in 2048, the speed for the program trained from the original TD learning was slightly faster since some overheads were incurred upon accessing multiple tables of feature weights for MS-TD learning.

| Reaching rate | 1-ply | 2-ply | 3-ply |
|---|---|---|---|
| 3072 | 0.08% (±0.06%) | 17.59% (±0.75%) | 38.90% (±0.96%) |
| 6144 | 0.00% (±0.00%) | 0.02% (±0.03%) | 0.45% (±0.13%) |
| Maximum score | 95,577 (±5,491) | 262,684 (±9,105) | 388,180 (±31,093) |
| Average score | 33,591 (±1,589) | 102,390 (±2,827) | 139,351 (±4,839) |
| Speed (moves/sec) | 153,132 (±700) | 16,107 (±331) | 359 (±1) |

Table 4: Results of 10,000 games for expectimax search using values learned from TD learning in Threes.

| Reaching rate | 1-ply | 2-ply | 3-ply |
|---|---|---|---|
| 3072 | 0.46% (±0.13%) | 47.03% (±0.98%) | 67.84% (±0.96%) |
| 6144 | 0.00% (±0.00%) | 1.40% (±0.23%) | 7.83% (±0.53%) |
| Maximum score | 135,069 (±10,880) | 518,348 (±28,006) | 700,181 (±13,817) |
| Average score | 33,808 (±1,687) | 157,025 (±5,272) | 220,393 (±12,393) |
| Speed (moves/sec) | 141,822 (±749) | 14,742 (±252) | 288 (±9) |

Table 5: Results of 10,000 games for expectimax search using values learned from 3-stage MS-TD learning in Threes.

### B. More Splitting Strategies

In this subsection, different splitting strategies for Threes are studied first. Strategy 1 adds one more splitting time at $T_{768}$ to the above simple 3-stage strategy. Fig. 20 and Fig. 21 (below) show its learning curves for the average scores and maximum scores, respectively. Strategy 2 adds one more splitting time at $T_{3072+1536}$, instead. Fig. 22 and Fig. 23 show its learning curves for the average scores and maximum scores, respectively. Table 6 (below) shows the experimental results of playing 10,000 games using 3-ply expectimax search for the above two strategies.

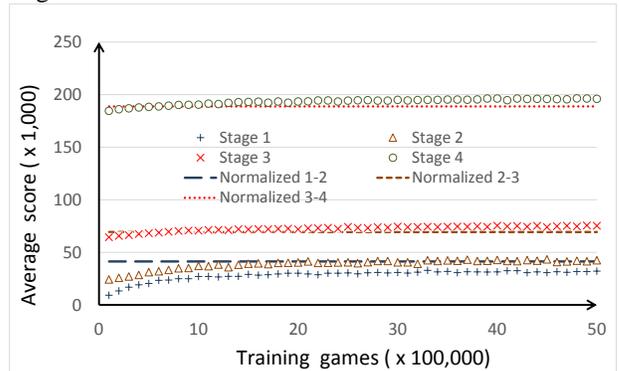

Fig. 20. Average scores in Threes for Strategy 1.

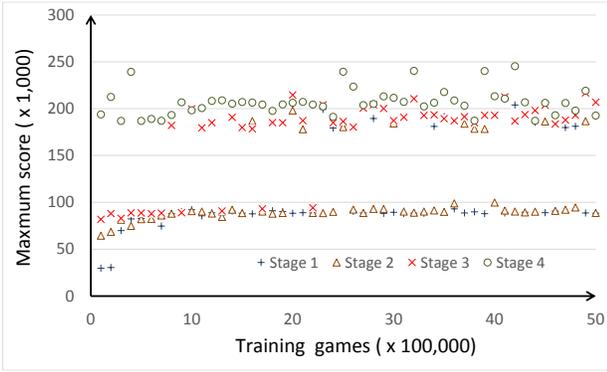

Fig. 21. Maximum scores in Threes for Strategy 1.

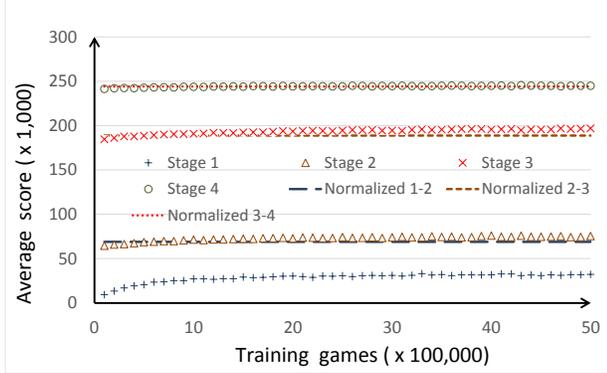

Fig. 22. Average scores in Threes for Strategy 2.

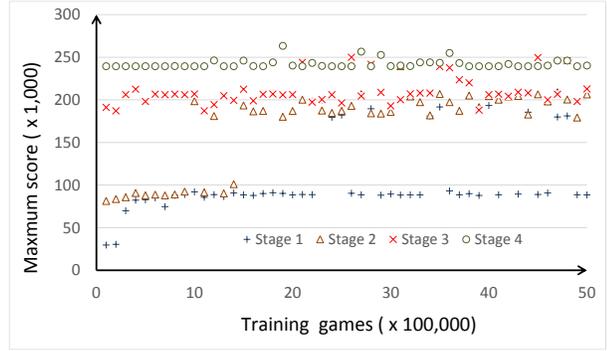

Fig. 23. Maximum scores in Threes for Strategy 2.

| Reaching rate | Strategy 1 | Strategy 2 |
|---|---|---|
| 3072 | 66.86% (±0.92%) | 67.84% (±0.92%) |
| 6144 | 6.29% (±0.48%) | 6.07% (±0.47%) |
| Maximum score | 665,494 (±13,606) | 664,485 (±12,194) |
| Average score | 213,059 (±8,537) | 214,277 (±8,450) |
| Speed (moves/sec) | 286 (±10) | 302 (±29) |

Table 6: Results of 10,000 games for 3-ply expectimax search using values learned from MS-TD learning with both Strategies 1 and 2 in Threes.

Fig. 20 to Fig. 23 demonstrate the improvements of MS-TD learning for the learning curves. However, when incorporated into expectimax search, these strategies did not perform better compared with the simple 3-stage strategy, especially in terms of 6144-tile reaching rates, shown in Table 6. This also showed that the benefit of splitting extra stages was diminishing, as observed in 2048. From above, the version with the simple 3-stage strategy performed the best among all the splitting strategies. This version is available openly at [7].

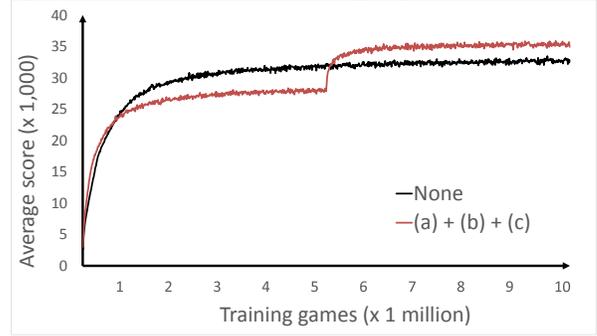

Fig. 24. Improvements for Threes TD learning with all of (a), (b), and (c) applied to the first stage.

We further applied the techniques (a), (b) and (c) in IV.E to Threes. Fig. 24 shows the improvement over un-tuned TD(0) training during the first stage. We extended this version with the simple 3-stage strategy, and obtained the following results in 10,000 games: a 6144-tile reaching rate of 8.91%, the average score 234,490, the maximum score 709,712, and a speed of 288 moves/sec.

## VI. CONCLUSIONS

This paper proposes multi-stage TD (MS-TD) learning, a kind of hierarchical reinforcement learning method, which improves the performance of 2048-like games effectively, especially for maximum scores and the reaching rates of large tiles. Our experiments in Sections IV and V demonstrated significant improvements when using MS-TD learning for both 2048 and Threes. For 2048, when 3-ply expectimax search was used, the 32768-tile reaching rate of using MS-TD learning with Strategy 3 was 18.31%, much higher than that for TD, which was 0%. After further improvements, our 2048 program reached 32768-tiles with a probability of 31.75% in 10,000 games and even reached a 65536-tile, the first ever reaching a 65536-tile to our knowledge. For Threes, when 3-ply expectimax search was used, the 6144-tile reaching rate of using MS-TD learning with the simple 3-stage strategy was 7.83%, much higher than that for TD, which was 0.45%.

Apparently, MS-TD learning can be easily applied to other 2048-like games. It is interesting and still open whether the MS-TD method can be combined with better variants of TD($\lambda$) learning, applied to other non-deterministic games and effectively used with expectimax search.

## ACKNOWLEDGMENT

This research was supported by NOVATEK Fellowship, Ministry of Science and Technology of the Republic of China (Taiwan) under the contract numbers MOST 104-2221-E-009-127-MY2 and 104-2221-E-009-074-MY2, and the National Center for High-performance Computing (NCHC) for computer time and facilities.

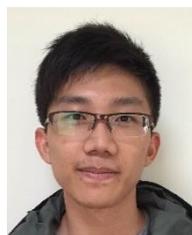

**Kun-Hao Yeh** is currently a master student in the Department of Computer Science at National Chiao Tung University. His research interests include artificial intelligence, machine learning and computer games.

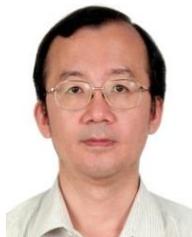

**I-Chen Wu** (M'05-SM'15) is with the Department of Computer Science, at National Chiao Tung University. He received his B.S. in Electronic Engineering from National Taiwan University (NTU), M.S. in Computer Science from NTU, and Ph.D. in Computer Science from Carnegie-Mellon University, in 1982, 1984 and 1993, respectively. He serves in the editorial board of the IEEE Transactions on Computational Intelligence and AI in Games and ICGA Journal. His research interests include artificial intelligence, machine learning, and volunteer computing.

Dr. Wu introduced the new game, Connect6, a kind of six-in-a-row game. Since then, Connect6 has become a tournament item in Computer Olympiad. He led a team developing various game playing programs, winning over 30 gold medals in international tournaments, including Computer Olympiad. He wrote over 100 papers, and served as chairs and committee in over 30 academic conferences and organizations, including the conference chair of IEEE CIG conference 2015.

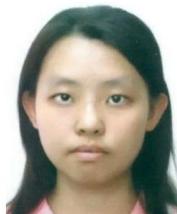

**Chu-Hsuan Hsueh** is currently a Ph.D candidate in the Department of Computer Science at National Chiao Tung University. Her research interests include artificial intelligence and computer games.

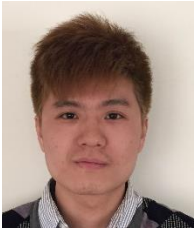
**Chia-Chuan Chang** is currently a master student in the Department of Computer Science at National Chiao Tung University. His research interests include artificial intelligence, computer games and grid computing.

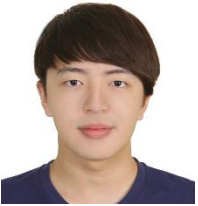
**Chao-Chin Liang** is currently a master student in the Department of Computer Science at National Chiao Tung University. His research interests include artificial intelligence, computer games and grid computing.

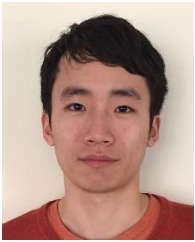
**Han Chiang** is currently a master student in the Department of Computer Science at National Chiao Tung University. His research interests include artificial intelligence and computer games.